# Textual Toxicity in Social Media: Understanding the Bangla Toxic Language Expressed in Facebook Comment

Mohammad Mamun Or Rashid[1]

**Abstract**

Warning: this paper contains text and image content that may be distressing or upsetting.

Social Media is a repository of digital literature including user-generated content. The users of social media are expressing their opinion with diverse mediums such as text, emojis, memes, and also through other visual and textual mediums. A major portion of these media elements could be treated as harmful to others and they are known by many words including Cyberbullying and Toxic Language[2]. The goal of this research paper is to analyze a curated and value-added dataset of toxic language titled ToxLex_bn[3]. It is an exhaustive wordlist that can be used as classifier material to detect toxicity in social media. The toxic language/script used by the Bengali community as cyberbullying, hate speech and moral policing became major trends in social media culture in Bangladesh and West Bengal. The toxicity became so high that the victims has to post as a counter or release explanation video for the haters. Most cases are pointed to women celebrity and their relation, dress, lifestyle are became trolled and toxicity flooded in comments boxes. Not only celebrity bashing but also hates occurred between Hindu Muslims, India-Bangladesh, Two opponents of 1971 and these are very common for virtual 'quarreling in the comment thread. Even many times facebook comment causes sue and legal matters in Bangladesh and thus it requires more study.  In this study, a Bangla toxic language dataset has been analyzed which was inputted by the user in Bengali script & language. For this, about 1968 unique bigrams or phrases as wordlists have been analyzed which are derived from 2207590 comments. It is assumed that this analysis will reinforce the detection of Bangla's toxic language used in social media and thus cure this virtual disease.

**1 Introduction**

The user-generated textual toxicity used in social media is a collective contribution that indicates the gigantic presence of intolerability in a virtual society. If the machine can recognize the toxicity with context and frequency then it is possible to regulator the harms. In the second generation of web technology especially web 2.0 (O'Reilly, 2007) new media, user can input their feedback against a website content, unwanted comments were filtered manually. But nowadays, toxic comments and feedback in social became enormous so this big data should be controlled (not directly filtered or removed entire comment) with some intelligent process ie keyword-based classifiers, distributional semantics, or deep learning classifiers (Salminen et al., 2020). These toxic data are not only immense but also have linguistical variation and complexity. So the current days the SOTA approach in detecting toxicity through data-driven approaches such as supervised machine learning or deep learning. After the successful development of the detection application, the social media user or the web admin can set a threshold of filterable inline toxicity. To do these, the major task is to define the responsible inline linguistic property which

---

could have carried out the toxicity in text, and then the second task is to suggest or remove/replace the word/phrase into a civil and acceptable word/phrase without hampering the meaning. To implement this strategy we should develop a toxic wordlist or phrase-list first, and the objective of this research paper is to develop a real-world toxic dataset with necessary features. So that an intelligent filtering system could be developed based on this dataset. Therefore, we have a main research question (RQ) for this study:

RQ 01: What are the quantitative features of the toxic words or bad words used in comments on social networks?

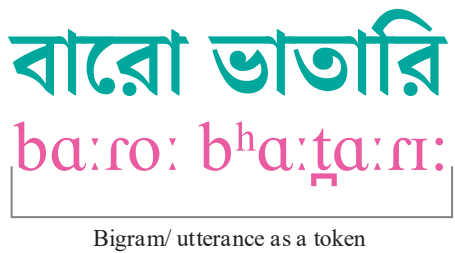
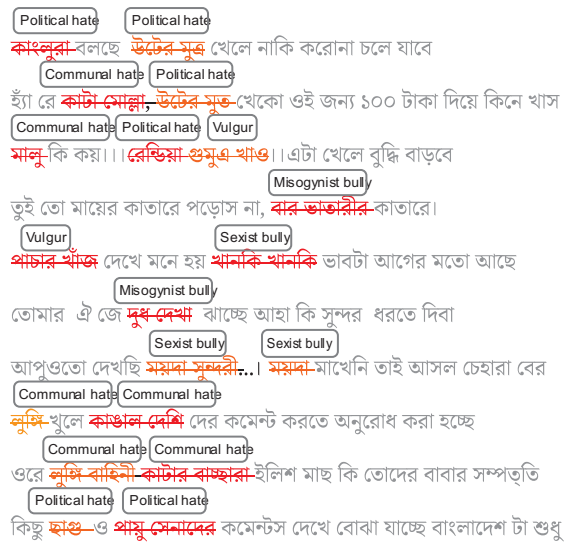

Figure 01: (a) Structure of this toxic lexicon dataset including bigram and other metadata (b) labeled bigram/s used in comments

Understanding a dataset is a pre-condition of any ML problem. Detecting toxic language is a real-life application of NLP as well as ML. In recent days, Deep learning methods are implemented where unannotated raw large datasets are used for detecting toxic language through multiclass classification where inlined in-text toxic phrases or words remain hidden and un unidentified. Therefore a multi-word or bigram lexicon could reinforce the system to detect the responsible specific part of the sentence and thus identify the toxicity in between a sentence. This approach will help to detect and filter the specific words rather than classify the entire sentence or document. Moreover, it will recognize the toxicity more precisely. To do this, there are some toxic wordlists for English but systematically coined exhaustive wordlists in English and even in other major languages very few. For Bangla/Bengali, we did not find a wordlist on cyber toxicity. Even, there is no study on what kind of cyber harm and hate speech Bengali users write in Bangla script on social media especially on Facebook.

**1.1 Previous Works**

Toxic language is a generalized term that aggregates all the negative textual expressions used on social media. All the terms covered under this generalized term can serve as synonyms, alternative keywords, or sub-categories. Under the concept of toxic language dozens of synonymous terms are usually penned by the scholars such as bad words (Kristiano & Ardi, 2018) (Sogg et al., 2018), cyberbullying (Fang et al., 2021), cyber harms (Agrafiotis et al., 2018), aggressive language (Kumar et al., 2021), abusive language (Lee et al., 2018), (Pitenis et al., 2020), impoliteness (Stoll et al., 2020), Hate speech (Tontodimamma et al., 2021), incivility (Sadeque et al., 2019), etc. All these terms are self-explanatory and have expressing negative concepts. Along with these general terms scholars have addressed some more toxic terms which are comparatively more specific or narrow and domain-dependent such as slang, profanity, humiliation, swear, personal attacks, provocation, threats, insults, etc.

Toxicity detection is a popular NLP problem so computational linguists and scientists published numerous papers on toxicity. But in terms of popular research trends, the development task of the toxic dataset from high resource language such as English to low-resourced language i.e Bangla is very few. Most research papers such as (Boudjani et al., 2020)(Pitenis et al., 2020)(Fang et al., 2021) have focused on toxicity recognition or detection. Recognition can be in two ways, the first is toxicity detection in the inline text, the second is to classify the entire short text as toxic/non-toxic, which behaves like spam filtering. Most papers follow the classification approach for toxicity detection (Kovács et al., 2021)(Fan et al., 2021). To do so, most papers publish the results of their models using datasets published by others. The main purpose of such papers is to show the performance of recognition or increase the existing performance.

There are also numerous research that creates datasets for themselves (Weld et al., 2021)(Saeed et al., 2021)(Gibert et al., 2018). Initiatives to create dedicated datasets carry different significance. Datasets of toxic languages can be of several types, such as some datasets have labels, some are non-annotated. In some datasets, each class is grouped, such as one group has toxic language and the other group has a non-toxic language. Some datasets contain only phrases or words responsible for toxicity. This approach can be called creating a dictionary or lexicon or word list.

English has been used more as a target language because it has more resource availability and more specific datasets. There is a lot of work in European languages like English (Bispo et al., 2019), French (Boudjani et al., 2020), Nordic (Coats, 2021), Turkish (Çöltekin, 2020), etc. However, in addition to these influential languages, there are research papers on subcontinental languages like Bengali (Romim et al., 2021) (Banik & Rahman, 2019), Urdu (Rizwan et al., 2020)(Sajid et al., 2020) (Saeed et al., 2021), and Hindi (Vashistha & Zubiaga, 2021)(Chopra et al., 2020), etc.

Some of the work that has been done on the toxic language used in social media has been done on certain media such as Twitter (Pitenis et al., 2020), Facebook (Romim et al., 2021), and YouTube (Salminen et al., 2020), Reddit (Salminen et al., 2020). Some studies have been done on specific domains such as Brexit(Fan et al., 2021).

Our current research is mainly to analyze a developed dataset having toxic words used in Facebook. In this dataset, the Bigram of Bangla language/script has been kept as a base entry. The main contribution of the current study is that it is a curated dataset and associates each entry with important features. For example, degree of toxicity, spelling standards, determination of features of base entries according to thematic classes, etc. This dataset also contains some quantitative metrics such as type-token frequency. Currently, the trend of toxicity dictation is towards extracting the desired language pattern using deep learning. However, it can be said without hesitation that the lexical approach or toxic lexicon will keep an important role in creating a real-life application for toxicity detection for Bangla/Bengali, which can be effective for other languages also.

## 2 Method

This paper is a quantitative analysis of TexLox_bn though it revealed the qualitative features of toxic words accordingly. As the dataset has been derived from social media comments, several manual steps such as data collection, cleaning, sorting, filtering, iteration, annotation, and value-addition tasks such as translation transcription have been followed. By completing these steps, a curated dataset called Toxic Lexicon (Bangla) or TexLox_bn has been created. The following steps are briefly described:

|   |   |
|---|---|
| a. | Collected 3200747 raw comments from 8 sources. |
| b. | Remained 2207590 comments in Bangla script after removing English characters and noise. |
| c. | Retrieved 3830555 bigrams from all Bangla comments, also 616719 unigrams. |

| | | |
|---|---|---|
| | d. | Filtered 375719 bigrams with the occurrences of more than 2 instances |
| | e. | Get 212377 bigrams as unique, deleting duplicates |
| | f. | Classified and labeled 2766 bigrams as toxic bigrams by 2-classes manual annotation |
| | g. | Remained 1968 bigrams and 1942 unigrams after removing stopwords from classified bigrams |
| | h. | Classified 1968 bigrams into 8 thematic classes of 1968 toxic bigrams |
| | i. | Search occurrence of 1968 toxic bigrams in 2207590 comments of 8 sources |
| | j. | Adding transcriptions, translation of 1959 toxic bigrams. |
| | k. | Adding 2-class label for spelling standard & 2-class label for the degree of toxicity |
| | l. | Export dataset with 9 column and 1959 rows as CSV |

Table 01: Dataset collection and processing steps in brief.

After the completion of the process, we found 1968 curated unique bigrams, and these bigrams have been written at least 104747 times in the comments. These bigrams have been enlisted by processing 3830555 bigrams and these processed bigrams have been collected from 3200747 Facebook comments. The source of the base bigrams is 08 Facebook pages; Of these, 5 are news media pages and the remaining three are women celebrity pages. Sources are mentioned in the datasets as BDH, PRM, PAL, MTH, KLK, DWB, BBC, ANB. The categories or classes of annotations that have been found are discussed below for some notable entries for each class on the dataset:

   a. Misogynist bully (Miso): These words are used to insult women. The high-frequency entries in this class are 'Baro Bhatari' (women has twelve partners: whore), 'Nichtola Bhara' (Renting ground Floor: slutting), 'Jamai bidesh' (Husband lives in broad), 'Chamra Bebsha' (Leather Business: prostitution), 'khat gorom' (Warming Bed), etc.
   b. Sexist & Patriarchic bully (Sexi): Sexual dissatisfaction, words that express perfection, these words express the power of men. The high-frequency bigram in this class is 'khankir chhele' (Son of a bitch', 'Sudanir Put' (son of a whore), 'Bal Pakna' (adultery attitude of an Immature), etc.
   c. Vulgar, incivility & sarcasm (Vulg): An ugly word used for ridicule, which cannot be used as formal. Some of the notable words in this category are 'Pa chata' (foot licker), 'kuttar baccha ' (son of bitch), 'jutar mala' (shoe garland), 'juta pita' (assault by shoe), 'ganjar gach' (cannabis plant), 'ganja kheye' (taking cannabis), etc.
   d. Political hate-words (Poli): A lot of aggressive words are used from political parties, ideologies, nationalism, patriotism. For example, 'gorur mut' (cow urine), 'uter mut' (camel urine), 'helmet bahini' (helmet force), 'kathal pata' (jackfruit leaf), 'gobor kheye' (taking cow dung), 'lungi bahinir' (lungi force: Bangladeshi), 'thala bati' (dish & bowl: Indian BJP ), 'kanglu ra' (kanglu ra: Bangladeshis), etc.
   e. Religion / communal hate words (Reli): Words that express fundamentalism or radicalism centered on religion or sectarianism. 'nastiker baccha' (Son of an Atheist), 'payu yoddha' (Anal Warrior), 'malauner baccha' (Son of Malaun), 'kath molla' (Conservative Mullah), etc.
   f. Racism on Body, Gender & Color (Raci): Bullying on skin color, texture, gender issues such as transgender, nonconform identity, and LGBT is used. Words related to body shaming 'transphobia and homophobia. ''pratibandhi gula' (Disables), 'kala naki' (what a dumb), 'pratibandhi naki' ( what a autistic), 'hijrar moto' (Like a transgender), etc.
   g. Moral Policing & Sewer (Mora): Giving policy advice to others according to one's own standards. Or the words expressed with curses and evil desires: 'parda karun' (cover yourself), 'hijab koi' (Where is the hijab), etc.
   h. Name trolling (Name): Distorting the name, adding new adjectives to the name to mock: 'goti danab' (Speed Monster), 'dalal bazar' (Broker Market), etc.

**3 Data Interpretation**

### 3.1 Toxic Lexicon: Derivatives of collective and virtual toxicity

This dataset made from user-generated content can be called lexicon or dictionary. This dictionary gives an idea of the collective toxicity that occurs in the virtual world. Since being used virtually collectively, these words have become a collective toxic language, giving Bengali users on social media an idea of the linguistic and scriptural culture. The various aspects of the collective attitude of the Bengali society on social media have been revealed in this dataset.

### 3.2 A Dictionary with linguistically diverse lemma

Each entry in this dictionary carries linguistic features differently, in particular, collective hate speech or sarcasm have been used in such a way that they are creating new meanings outside of conventional meaning. That is, the words on the vocabulary list have added new semantic meaning. Analyzing some words, it is seen that there is a difference between denotative meaning and connotative meaning. For example, the dictionary meaning of words like are 'Baro bhatari' (women has twelve partners: whore), 'Nichtola Bhara' (Renting ground Floor: slutting), 'Chamra Bebsha' (Leather Business: prostitution),
'moyda suji' (flour semolina: beauty through cosmetics), 'moydar bosta' (flour bag: beauty using extreme cosmatics), 'multi-plug' (connected with multiple line: whore), 'rate koto' (what is your rate: rate for sex), etc. does not apply here. The connotative and social meaning of the word has to be emphasized here.

An analysis of this wordlist has also revealed a different technique for users to express toxicity, where one or more letters of a word are slightly altered, such as 'motherboard', 'sudani', 'khan ki', 'mother toast'. , 'tode na' is a slightly changed word from obscene words that are considered offensive by community standards. The reason for this change may be the avoidance of the obligation to express obscenity in the user or author's strategy.
Some of the words in the toxic dataset are highly context-dependent. However, most of these words have become very popular in the Bengali commentator community. These are also distant from denotative meaning. These words refer to another person, object, class, or idea. For example, 'goti danob' (speed monster) for a cricket official of Bangladesh , 'bhaipo' (nephew) addressing a political person of West Bengla, 'Chawala' (tea man) indicating politician of India, 'kathal pata' (jackfruit leaf) indicating supporter of a specific ideology found in Bangladesh, 'paat khete' (jute field) indicating an easily hidden place for rape, etc.

### 3.3 Included spelling variants to cover a wide range of toxic words

Basewords are usually unigram in traditional toxic wordlists or bad word dictionaries. But if Unigram is counted as a dictionary entry, the range of toxicity recognition will become narrow. This dataset is publishing the baseword from the user's text, so all the spelling variations including misplaced words and typos are included in the dictionary. Analyzing the dataset shows that the spelling variant should be an important feature of the dataset and so it has been done. Standard spelling is usually followed in edited text, but since this dataset has been collected from the public, it contains heterogeneous spelling variants, but some word spelling variants are numerous. For example, there are 06 variants of the word 'camel's mouth' and there are 06 variants of the word 'barovatari' also. If the standard spelling is kept in the dictionary then the accuracy of the variants will be affected during recognition. The variants of the words 'camel's mouth' and 'barovatari' are: 'outer mue' (উটের মুএ) 15, 'outer mut' (উটের মুত) 276, 'outer mutra' (উটের মুত্র) 86, 'outer muut' (উটের মূত) 20, 'outer mutra' (উটের মূত্র) 230, 'uther mut' (উঠের মুত) 27; 'bar batari' (বার বাতারি) 70, 'bar batari' (বার বাতারী) 12, 'baro bhatari' (বারো ভাতাড়ি) 6, 'baro bhatari' (বারো ভাতারি) 459, 'baro bhatari' (বারো ভাতাড়ী) 6, 'baro bhatari' (বারো ভাতারী) 102 etc.

### 3.4 Extensive coverage with collocated neighbors of Toxic word

A unigram alone can be a toxic word. But if any other word is added after that, it becomes a bigram, then it contains a lot of contexts; it may become a regular word or a toxic word. This means it is relatively possible to recognize the features of a language through a bigram. Analyzing this dataset, it has been found that the word 'juta'/shoe (জুতা) has a collocation of 08, the word 'moyda'/ flour (ময়দা) has a collocation of 09 and the word '

kapor'/ cloth (কাপড়) has a collocation of 06 (Table 02). In other words, if the dataset was created through Unigram, the toxic word could not be recognized with this collocation, that means these bigrams are useful to define many contexts. Also, the domain space has been extended as this dataset has been created based on Collocated Neighbor.

| Exp 1 | IPA | Occ. | Exp 2 | IPA | Occ. | Exp 1 | IPA | Occ. |
|---|---|---|---|---|---|---|---|---|
| জুতা চুর | d͡ʒuṯa: t͡ʃurə | 32 | ময়দা মাখা | məẏəda: ma:kʰa: | 14 | কাপড় খুলে | ka:pəɾə kʰule: | 112 |
| জুতা দিয়ে | d͡ʒuṯa: dɪẏe: | 103 | ময়দা মাখে | məẏəda: ma:kʰe: | 5 | কাপড় ছাড়া | ka:pəɾə t͡ʃʰa:ɾa: | 92 |
| জুতা দিয়া | d͡ʒuṯa: dɪẏa: | 238 | ময়দা সুন্দরী | məẏəda: sun̪d̪əri: | 39 | কাপড় ছাড়াই | ka:pəɾə t͡ʃʰa:ɾa:ɪ | 20 |
| জুতা পিটা | d͡ʒuṯa: pɪṯa: | 306 | ময়দা কম | məẏəda: kəmə | 13 | কাপড় ছোট | ka:pəɾə t͡ʃʰo:ʈə | 16 |
| জুতা পেটা | d͡ʒuṯa: pe:ṯa: | 279 | ময়দা বেশি | məẏəda: be:ʃɪ | 7 | কাপড় তুলে | ka:pəɾə t̪ule: | 18 |
| জুতা মার | d͡ʒuṯa: ma:rə | 383 | ময়দা মাখলে | məẏəda: ma:kʰəle: | 8 | কাপড় নাই | ka:pəɾə n̪a:ɪ | 21 |
| জুতা মারা | d͡ʒuṯa: ma:ra: | 208 | ময়দা মেখে | məẏəda: me:kʰe: | 36 | কাপড় নেই | ka:pəɾə n̪e:ɪ | 40 |
| জুতার বাড়ি | d͡ʒuṯa:rə ba:ɽɪ | 208 | ময়দা সুজি | məẏəda: sud͡ʒɪ | 41 | কাপড়চোপড় খোলা | ka:pəɾət͡ʃo:pəɾə kʰo:la: | 28 |
| জুতার মালা | d͡ʒuṯa:rə ma:la: | 309 | ময়দা সুজির | məẏəda: sud͡ʒɪrə | 16 | কাপড়ের অভাব | ka:pəɾe:rə əbʰa:bə | 24 |

Table 02: Dataset covers bigrams to understand the context of Bangla toxic words

## 4. Discussion
### 4.1 Misginyst and sexual bad words are highly dense and frequent
 After the completion of toxic words analysis, it can be accredited that women are the direct or indirect target of toxicity and the reason for the expression is the unsatisfied sexual desire of the patriarchal society, which is expressed using various sexual and misogynistic terms. In the collected wordlist, misogynist and sexist words are more in number and more in terms of frequency. The trajectories of Figure 02 are showing the comparative Type-Token percentage of this dataset. The dataset also shows that the two main classes, the misogynist bully and the sexist and patriarchal bully, account for 19 percent and 30 percent of the total word count, respectively. And their frequencies are 19 and 26 percent, respectively. This is followed by the 'Vulgar, Invisibility and Circumcision' classes. Regardless of the issue of posts or news in Facebook comments or user-generated content, there is always a tendency to create sarcasm. No matter how rude and inaudible this sarcasm is, it is being written. The type of this class, the higher the percentage presence of tokens, are 30 and 23, respectively.
 Apart from these three classes, another type of topic is present the toxic wordlist, which is a political or ideological conflict. The cause of this conflict may be religion, country, nationality, or community. The radical or fundamental position on the identity or ideology is such that such comments can be identified as hate speech. This class is also strongly increasing the toxicity of social media. The communal hate-speech or hate-word is divided into two classes in this dataset, one political, related to power or nationality; The other is related to religion, belief, or cultural group. The type of political-hate-words in the dataset, the presence of both tokens is 14%. Of religion or communal hate-words. Although the frequency is 04%, their variety is more than 07%. The rest of the classes in this dataset are minor and their frequency and population are below 1%.

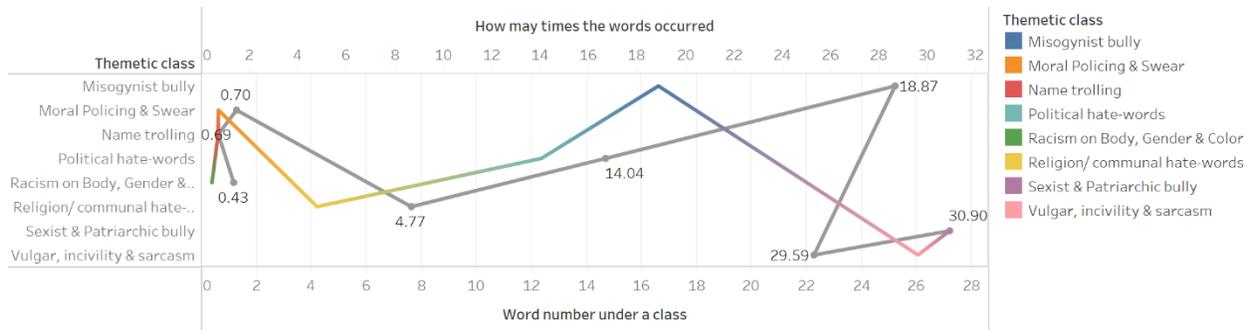

Figure 02: Type token distribution of Bangla toxic language in Facebook

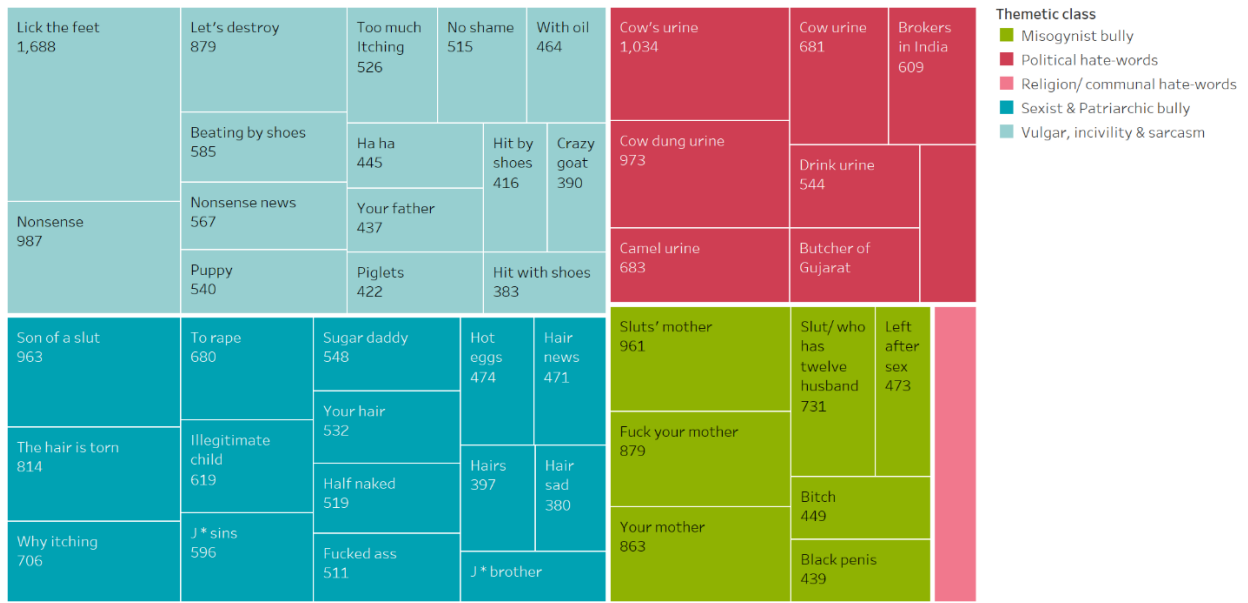

Figure 03: Misginyst and sexual bad words are highly dense and frequent in this dataset.

This numerical analysis proves that Bengali social media users spark on feminism and religion and power-centric ideologies. And most of the time this flare-up manifests itself through incivility. In other words, Bengali users are very interested in sexual issues and communal ideology (Figure 03). So there is a high traffic of news or posts about women and religion. Their position has been revealed not only in terms of numbers but also in terms of expression. Often trolls and circuses use informal language to express cyber-bullying in a way that can easily be called toxic.

**4.2 Toxic word classes are biased with sources**

There is a dependency on the category of cyber-bullying with the type of social media page, which is being proved easily due to the statistically positive correlation. Our study found that the pages of three female celebrities had a predominance of misogynistic and sexist words. And the number of politically toxic tokens on these pages is less than one thousand, so there is no political token here, on the other hand, the news media pages contain a large number of tokens in almost all 5 main classes. These pages have the presence of political and communal hate speech with misogynist and sexist bad words. Figure 04 and Figure 05 are showing that misogynistic bullying and sexist bad words are more prevalent on the pages of female celebrities and political and other classes are more prevalent in news media pages.

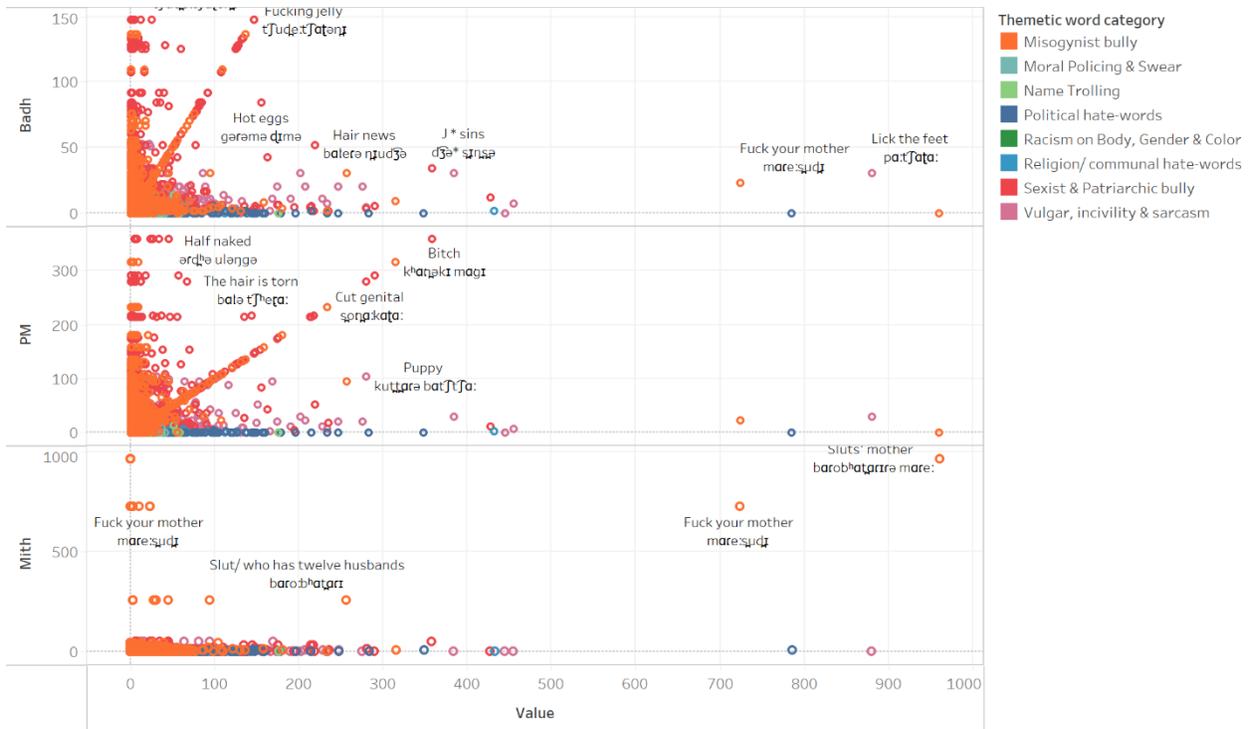

Figure 04: Toxic words correlate with source pages such as the pages of women celebrities who receive misogynist and similar types of comments.

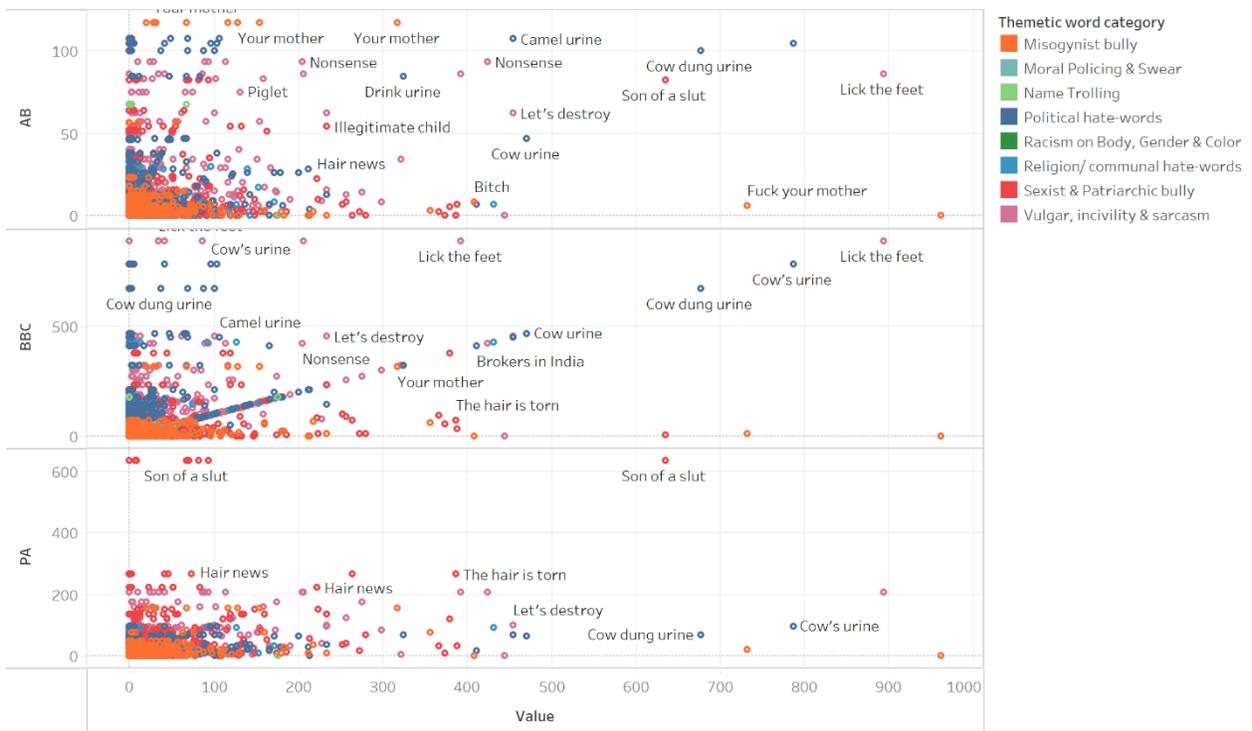

Figure 05: Both political hates and sexist bullies are frequent in the newspaper page comments.

### 4.3 Usecases of the toxic lexicon: Recognition models, Filtering

Toxic Lexicon has several use-cases, among the most simple and useful use-case will be toxic text filters on any social media such as Facebook, YouTube, Twitter, WordPress, etc. WordPress currently uses short vocabulary sets for different languages through some plugins, which run through a static approach such as the dictionary lookup technique. As there is no dictionary or exhaustive all-inclusive slang words in the Bengali language used by social media users. At the same time, there is no categorization of the Bangla toxic word list. Therefore, this lexicon will be effective for filtering these toxic words. However, The subtle use case of this lexicon is to help create machine learning models. This dataset will help in preparing the dataset, especially for toxicity recognition. This dataset can be used as a gold standard data that will not only filter out spam detection but also detect which category of language has been filtered. This will allow you to recognize the use of disrespectful offensive words or phrases in any document, text, or literature, including comments.

### 4.4 Limitations

Although this dataset has been prepared following the proper procedure, some limitations need to be mentioned. The diagrams of these datasets at the time of annotation are divided into different thematic classes but it is quite difficult to distinguish between these thematic classes. For example, misogyny and sexism cannot be distinguished, as the victims or targets of these two classes are mainly women. Similarly, it is difficult for an annotator to distinguish which class some of the words in a hate speech, such as 'kanglu' (Bangladeshi), 'mutkhor' (urine eater), 'kata' (circumciser/ Muslim), 'akata' (non-circumciser/ Hindu), etc., belong to, as they can represent almost two synonymous classes. These words are usually exchanged between Indian and Bangladeshi people, which may be due to their religious affiliation, sectarianism, or their attempt to promote their political identity. In the Bangladesh-India context, sometimes the boundary between religion-centric and political hate speech cannot be identified.

This dataset does not include a large portion of the original dataset. As many users write Bangla in English script, which is not in this dataset. This dataset only discusses toxicity written in Bengali. The frequency threshold of Bigram has also been set to 'More Than Two', omitting many toxic words. Although no word used twice in a few million comments should be included in the Corpus Driven wordlist.

The dataset is anonymous due to copyright policy and the source sentence could not be linked because it is only Facebook's data. Which could have further enriched this dataset.

This dataset is not developed through multi-human annotations, so it does not have an inter-annotator agreement. As a result, the annotation of the dataset may have the subjective bias of the researcher.

### 5 Conclusion: Future of cyber harms and safe internet

This current dataset will help the community to reduce cyberbullying and other cyber harms on publicly open pages. This is because social network platforms will be able to automatically filter out different categories of bad words or notify the commenter via alert notice for reconsideration. This will ensure Extreme Hate Speech and Badword Free Safe Internet. Also, this dataset will be a sample toxic dataset and will work as a template for all neighboring low-resourced languages of Bangla including all Indic languages.

### 6 Misc.
### 6.1 Ethical aspects

This study follows the ethical policy set out in the corresponding publishing organization's code of Ethics and addresses the ethical impact of presenting a new dataset.

This dataset protects the privacy of individuals. No one's personal information was taken during the data collection, moreover, no records were kept. It uses user-generated data, the names and identities of the users are not here, sometimes the '*' mark has been used to blur the identity. Who said, and whom said - both specific user identities are not kept here.

Also, there is no full sentence or phrase here. Only the utility unit of the comment, which is responsible for toxicity, has been extracted as a bigram. Words have minimal context as a result of using a bigram. This feature has made every bigram universal. As a result, this dataset can be published without any restrictions.

### 6.2 Data availability statement

The raw data such as user-generated short text collected for this research cannot be freely distributed due to facebook's terms and conditions. However, the bigram data are distributable as they are context-free and anonymous. The curated value-added dataset of this is available at Data set]. Mendeley. https://doi.org/10.17632/9PZ8SSMC49.1 [now attatched].

### 6.3 License

Open Access This article is licensed under a Creative Commons Attribution 4.0 International License. To view a copy of this license, visit http://creativecommons.org/licenses/by/4.0/
6.4 Disclosure statement
No financial interest or benefit has arisen from the research.

### 6.5 Conflict of Interest

The author declares that they have no conflict of interest.